\documentclass{article}
\usepackage{spconf,amsmath,graphicx}

\usepackage{amsfonts}
\usepackage{arydshln}
\usepackage{multicol}
\usepackage{multirow}
\usepackage{booktabs}
\usepackage{subcaption}

\def\AUC{{\mathrm{AUC}}}
\def\ECE{{\mathrm{ECE}}}
\def\NCE{{\mathrm{NCE}}}
\def\NPV{{\mathrm{NPV}}}
\def\TNR{{\mathrm{TNR}}}

\def\NT{{\mathrm{NT}}}
\def\PRC{{\mathrm{PRC}}}
\def\ROC{{\mathrm{ROC}}}
\def\WCR{{\mathrm{WCR}}}

\def\RMSE{{\mathrm{RMSE}}}
\def\truelabel{{\mathrm{True}}}

\title{Word-level confidence estimation for RNN transducers}

\name{Mingqiu Wang, Hagen Soltau, Laurent El Shafey, Izhak Shafran}

\address{Google Research \\
{\small {\sffamily \{mingqiuwang,soltau,shafey,izhak\}@google.com}}}

\begin{document}
%\ninept
%
\maketitle
\begin{abstract}
Confidence estimate is an often requested feature in applications such as medical transcription where errors can impact patient care and the confidence estimate could be used to alert medical professionals to verify potential errors in recognition. 

In this paper, we present a lightweight neural confidence model tailored for Automatic Speech Recognition (ASR) system with Recurrent Neural Network Transducers (RNN-T). Compared to other existing approaches, our model utilizes: (a) the time information associated with recognized words, which reduces the computational complexity, and (b) a simple and elegant trick for mapping between sub-word and word sequences. The mapping addresses the non-unique tokenization and token deletion problems while amplifying differences between confusable words. 
Through extensive empirical evaluations on two different long-form test sets, we demonstrate that the model achieves a performance of 0.4 Normalized Cross Entropy ($\NCE$) and 0.05 Expected Calibration Error ($\ECE$). It is robust across different ASR configurations, including target types (graphemes vs.~morphemes), traffic conditions (streaming vs.~non-streaming), and encoder types. We further discuss the importance of evaluation metrics to reflect practical applications and highlight the need for further work in improving Area Under the Curve ($\AUC$) for Negative Precision Rate ($\NPV$) and True Negative Rate ($\TNR$).

\end{abstract}
\begin{keywords}
Confidence estimation, RNN-T, ASR
\end{keywords}
\section{Introduction}
\label{sec:intro}
Automatic speech recognition (ASR) is typically the first step in understanding spoken content. Even for a high-performance ASR system, the misrecognition between similar sounding medications, for example, could have a disproportionately high impact on downstream processes. Confidence estimates on the correctness of the recognized words can potentially safeguard against propagating high-risk errors. They can also be used to prompt users for additional confirmation of the veracity of the content in specific audio segments, a task which can otherwise be onerous in long conversations.

% Traditional ASR model and confidence.
There has been more than two decades of research in developing confidence estimates for ASR outputs~\cite{Siu1999,Kemp1997}. In conventional ASR systems, the probabilities of decoding lattices or n-best lists are typically used to directly compute confidence scores~\cite{wessel1998using}. Confidence scores are often computed by estimating word posterior probabilities from lattices~\cite{evermann2000large} or word confusion networks~\cite{mangu2000finding}. A more flexible approach involves predicting confidence scores using a separate classifier~\cite{Kemp1997} whose input consists of a variety of acoustic and lexical features extracted from the ASR system that include lattice posteriors, acoustic/lexical latent representations, phonetics, and word duration, etc~\cite{seigel2011combining, kalgaonkar2015estimating, ragni2018confidence, li2019bi}.

% Current RNN-T model is over confident, and why they are over-confident?
Neural based end-to-end ASR systems such as recurrent neural network transducers (RNN-T)~\cite{Graves2012} have recently become popular and they have achieved state-of-the-art word error rate (WER) for a vast majority of speech recognition tasks. Unlike conventional ASR systems where the acoustic and the language models are trained separately, in these end-to-end models both components are jointly optimized using a single cost function such as likelihood. While they improve performance substantially, these models tend to exhibit very skewed posterior probabilities~\cite{ovadia2019can, Guo2017}. Several factors lead to the skewed posteriors: 1) Neural networks in general have powerful representation ability and can overfit the training data. 2) In neural network models optimized by negative log-likelihood, the likelihood can be increased further by sharpening the logits even if the predictions of the logits are already correct~\cite{muller2019does}, a phenomenon that can be observed by tracking the posteriors as the training continues after the convergence of the ASR model. 3) RNN-T has an integrated language model (LM) trained using reference transcripts (teacher-forcing), which is unlike connectionist temporal classification (CTC) models~\cite{Graves2006b}.
% Compare to existing works. 
% Temperature scaling, etc. David and Qiujia's paper. 

There has been a surge of recent related work often categorized under "calibration" or "reliable deep learning". One simple and popular approach uses temperature scaling, a variant of Platt scaling, on the logits to flatten the posteriors~\cite{Guo2017}.  A more sophisticated approach that has proved to be very effective utilizes ensembles of models~\cite{lakshminarayanan2016simple}, however this approach is computationally too expensive for ASR systems. Lately, neural confidence models have been proposed where the target is constrained to "correct" and "incorrect". This approach has the flexibility to use various features from the ASR models~\cite{li2020confidence, qiu2021learning, li2021residual, qiu2021multi, woodward2020confidence}. In one such approach, a confidence estimation model is integrated with a Listen-Attend-Spell (LAS) model and the word-level scores are obtained by averaging the sub-word predictions associated with the correct/incorrect targets~\cite{li2020confidence, woodward2020confidence}. The mapping from sub-word to word sequences are not unique and when most sub-words are recognized correctly in a misrecognized word, averaging the sub-word predictions results in weak learning signal and this could be problematic in cases where hypothesis has sub-word deletions. This was addressed by an auxillary loss~\cite{qiu2021learning,qiu2021multi}. 

The work reported in this paper differs from other previous work in the following respects. 
First, our goal is to provide confidence scores for long-form streaming rich transcription which reflects real-world applications, for example, in medical transcription where the audio input could be half to an hour long. In contrast, previous works are in non-streaming conditions with short utterances and without speaker labels, punctuation or capitalizations.
Second, to the best of our knowledge, the proposed confidence model is the first one tailored for RNN-T models. By utilizing the emission times, we demonstrate that the model effectively extracts important acoustic information from the ASR encoder, and reduces the computational complexity.
Third, we addressed the non-unique tokenization and token deletion problems with a simple and elegant trick. Our method simply modifies the training targets without changing the model architecture. Besides, this trick also amplifies the error signals and improves performance.
Fourth, unlike previous work where the empirical evaluations are only performed on a single ASR system, we perform comprehensive evaluations on eight ASR settings: grapheme vs.~morpheme, streaming vs.~non-streaming ASR, and different ASR model architectures. We demonstrate that our model is robust and generalizes across all the configurations. 
Fifth, we highlight the importance of evaluation metrics to reflect the real-world application of post-ASR editing as in medical transcription.

\section{Model}
\subsection{Model Architecture}

Our proposed confidence model predicts confidence score for each sub-word unit (a.k.a., token) in the hypothesized sequence using both acoustic and lexical features associated with the unit.

\noindent \textbf{Features} \label{sec:model:features}: 
In our model, the \emph{acoustic feature} for each unit is generated from a sub-sequence in the ASR encoder output centered around the time when the unit was predicted by the RNN-T model (emission time). The length of the sub-sequence is a hyper-parameter ($k$ as defined below). The \emph{lexical feature} for each sub-word unit is the token's embedding vector, which could be either trained from scratch, or initiated from the RNN-T model's token embedding. 

To generate acoustic feature for a sub-word unit, we extract $k$ encoder outputs to the left and right of the sub-word emission time. This $2k+1$ long sub-sequence is mapped into a single vector as the acoustic feature. We investigated different methods for the mapping including attention or direct concatenation. When using attention, we used the lexical feature vector as a query to compute cross-attention with $2k+1$ long sub-sequence.
The acoustic and lexical features are concatenated to form the input feature for this sub-word unit.

\begin{figure}[h]
    \vspace{-.1in}
    \centering
    \includegraphics[width=0.95\columnwidth]{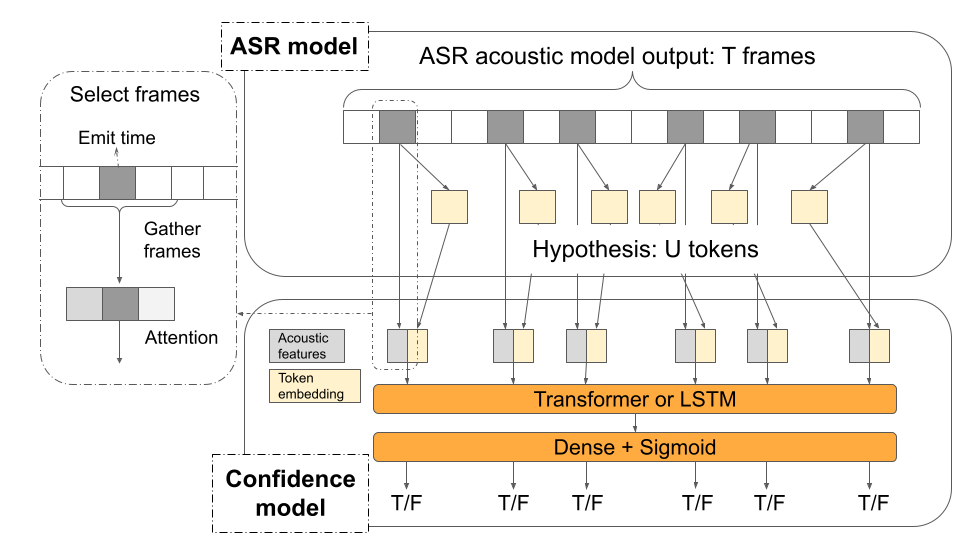}
    \caption{Confidence model architecture.}
    \label{fig:confidence_model}
    \vspace{-.1in}
\end{figure}

\noindent \textbf{Model Structure}:
The model consists of a sequence layer followed by a dense layer, as illustrated in Figure~\ref{fig:confidence_model}. We investigated two options for the sequence layer -- a long-short term memory (LSTM) and a transformer. Since typical transformer's compute attention over the entire span and that can be computationally expensive in long-form audio, we adopted a variant where the attention is limited to a fixed window (a hyper-parameter) and the receptive field is build up hierarchically with the number of layers~\cite{Povey2020, Qian2020}. The output from the sequence layer is fed to a dense layer and finally passed through a sigmoid function to generate a confidence score. The model is trained using a cross-entropy loss. 

Our lightweight model has approximately $4$ million parameters and incurs computation cost of $O(U)$ in the length $U$ of the sub-word units which allows us to scale the model easily for audio with long duration. Note, in previous work that did not utilize the time information, the computation cost incurred was $O(UT)$ due to the cross attention with the whole ASR encoder outputs, where $T$ is the length of audio frames.

\subsection{From Sub-Word to Word-Level Confidences}
Recall, our confidence model predicts correctness of each sub-word unit
% independently 
as in a classification task. As such, for training the model, apart from the windowed encoder output and the sub-word embedding vector, we need the target class indicating the correctness of the prediction. 

One way to identify the correctness of the sub-word is to map the reference word sequence to the corresponding sub-word sequence, compute the Levenshtein distance between the two sub-word sequences and identify the sub-word units in error. Since the correctness can only be associated with sub-words that are hypothesized by the ASR system, it does not account for words that are incorrect because of a missing sub-word unit. For example in Table~\ref{alignment}, all the hypothesized sub-word units in the two word sequence {\it son // g} that corresponds to the reference word {\it song} will be marked correct even though the hypothesized words are in error.

An alternate method that overcomes this issue first aligns the hypothesized words with the reference words, then transfers the correctness to the last sub-word unit in each hypothesized word~\cite{qiu2021learning}. As the example in Table~\ref{alignment} illustrates the correctness of the two hypothesized words {\it lov // ely} is transferred to the last sub-word units {\it lo // ve // ly}. At inference time, the confidence scores are masked for all the sub-word units except for the last one in the word.

We apply a small trick to this mapping which turns out to be effective. Instead of applying the correctness only to the last sub-word units, we apply them to all the sub-word units in the incorrect word, as shown in Table~\ref{alignment}. Since most ASR systems have substantially fewer incorrect words compared to correct words, this trick allows us to increase the number of negative examples. For example, this increases the total number of training examples by a factor of two for morphemes and a factor of four for graphemes. Additionally, this mapping gives due credit to different segmentations of correctly recognized words (e.g., lov-ely vs lo-ve-ly). At inference time, we generate confidence scores at word-level by averaging the predicted sub-word level scores, which has a simpler implementation than the masking approach.

\begin{table}[ht]
\vspace{-.1in}
\centering
\begin{tabular}{lccccc}
\toprule
Hypothesis & \_lov & - & ely & \_son & \\ 
Reference & \_lo & ve & ly & \_son & g  \\ 
\midrule
\multicolumn{6}{c}{Corresponding targets for confidence model} \\
\hdashline
Morpheme & False & False & False & True & False \\ 
Word \cite{qiu2021learning} & - & - & True & - & False  \\ 
Word (ours) & True & True & True & False & False  \\ 
\bottomrule
\end{tabular}
\caption{Illustration of mapping targets at the token and word levels. Start of words are denoted by \_ symbols.}
\label{alignment}
\vspace{-.3in}
\end{table}

\section{Experimental Setup}
\subsection{Datasets} 
\label{sec:datasets}
We performed empirical evaluations on a medical dictation and a medical conversation corpora~\cite{shafran-etal-2020-medical}.

\emph{The medical dictation corpus} consists of $\approx 5$K hours of audio recordings along with their ($\approx 100$K) transcribed clinical notes. The corpus contains $4$K unique speakers of which about 30\% were female speakers. The dictations spanned multiple specialities including radiology, internal medicine, family medicine, cardiology, psychiatry and oncology. The corpus was split into $\approx 4$K hours of training data, $\approx 400$hrs of held-out data, and $\approx 56$hrs of test data, with no speaker overlap between the three sets. The transcripts contain clinical abbreviations (e.g., a1c), clinical measures (e.g., blood pressure of 120/70), special annotations to identify spoken commands for formatting such as period, comma, capitalization, next paragraph, and section headings.

\emph{The medical conversation corpus} is significantly larger and consists of about $100$K ($\approx 15$K hours) manually transcribed audio recordings of clinical conversations between physicians and patients from a wide range of specialities. The turns are attributed to the speakers in the conversation and are categorized into four roles -- providers, patients, caregivers and others. Additionally, the transcripts includes punctuation and capitalization, similar to the dictation transcripts. Test set consisted of 500 conversations where the providers don't overlap with the training set (we don't have information -- anonymized identity -- on the patients or caregivers). For more details, see~\cite{shafran-etal-2020-medical}.

\subsection{ASR system}

Our ASR system is based on an RNN-T architecture with a uni-directional transformer encoder and an LSTM decoder~\cite{soltau2021medical}. Specifically, the encoder consists of $15$ transformer layers with a model dimension of $1,024$ and $8$ attention heads, while the decoder consists of two LSTM layers with $1,024$ units. As mentioned before, we use a variant of transformer layer with limited attention of fixed window size~\cite{Povey2020, Qian2020}. The model vocabulary is based on graphemes and includes lower-case and upper-case letters, digits, punctuation and speaker role symbols, resulting in a total vocabulary size of $102$ tokens. 

We investigated variations from the primary ASR system, including LSTM-based encoders, non-streaming condition with bi-directional encoders, and morpheme-based vocabulary~\cite{Virpioja2013}. The influence of these settings on confidence performances is described in Sec~\ref{result:different_asr}.

Our primary ASR model achieves a word error rate (WER) of $21.1\%$ on the conversation development set and $10.8\%$ on the dictation development set. The corresponding confidence models are evaluated on entire audio utterances, which are segmented automatically using a neural network-based speech detector to mimic the traffic in a production environment.

\subsection{Evaluation Metrics}
We rely on several metrics to evaluate the performances of our confidence models.

\noindent \textbf{Normalized Cross Entropy ($\NCE$)} measures the relative decrease in uncertainty brought by the conﬁdence estimation about the correctness of a word~\cite{nce2017}. This metric ranges from $-\infty$ to $1$, with higher $\NCE$ scores corresponding to better confidence estimates. When the conﬁdence model performs worse than the chance (setting confidence score to average word correctness ratio), the $\NCE$ score is negative. The metric takes into account the performance of the associated ASR system and achieves its highest value when all correct and incorrect words are assigned $1.0$ and $0.0$ respectively.

\noindent \textbf{Expected Calibration Error ($\ECE$)} represents the difference in expectation between confidence and accuracy~\cite{naeini2015obtaining}. $\ECE$ achieves the best performance (score of $0.0$) when the confidence scores matches the measured correctness of words with that score. The lower $\ECE$ scores correspond to the better confidence estimation. 

\noindent \textbf{Area Under Curve ($\AUC$)} In addition to the commonly used $\AUC_{\ROC}$ and $\AUC_{\PRC}$, we also show $\mathrm{AUC_{NT}}$, the area under the $\NPV\sim\TNR$ curve (Negative Predictive Value vs. True Negative Rate). In practical applications, users or downstream applications wish to pick a confidence threshold to identify misrecognized words. Instead of using $\AUC_{\ROC}$ and $\AUC_{\PRC}$ which are dominated by correct predictions, the most pertinent performance trade-off curve for picking such a threshold is $\NPV$ (similar to negative recall) vs. $\TNR$ (similar to negative precision). $\AUC$s are numbers between $0$ and $1$ and higher values indicate more reliable confidence scores.

A few points of caution. The choice of metrics is important in characterizing the confidence scores. Two of the metrics -- $\AUC_{\ROC}$ and $\AUC_{\PRC}$ -- are influenced disproportionately by correct words and paint an overly optimistic picture of performance. For medical application, $\NCE$ and $\AUC_{\NT}$ are more relevant for practical applications. As an aside, two confidence estimators can have the same $\AUC$ value but different $\NCE$ values~\cite{li2020confidence}. We do not include the utterance-level evaluation metrics such as $\RMSE$, since our application requires word-level confidence for long-from audio with duration ranging between 15 minutes to an hour.

\section{Experimental Results}
We categorized our experiments into two groups -- modification of RNN-T posteriors and a dedicated confidence model.

\subsection {Modified Posteriors as Confidence Scores}
We evaluated the simple approach of estimating the confidence scores by simple modifications of posteriors and the results are reported in the upper portion of Table~\ref{confidence:dictation_model}. The performance is measured over all the recognized events including case-sensitive words and punctuation. The word-level scores were computed by averaging the estimates over all the sub-word units, as mentioned earlier.

When the posteriors from softmax of RNN-T's joint layer are directly used as confidence scores, the $\NCE$ score is $0.01$ which is close to chance ($\NCE = 0.0$). This is not surprising since the RNN-T posteriors are known to be skewed. A simple method of improving confidence scores is to construct a pairwise mapping (look-up table) from the posteriors to the confidence scores~\cite{evermann2000large}. This nudged the performance to an $\NCE$ score of $0.07$, without any impact on $\AUC_{\NT}$. Next, we applied temperature scaling~\cite{Guo2017} with $T = 2.0$ to achieve a small improvement with $\NCE = 0.7$ and $\AUC_{\NT} = 0.28$.

We investigated the impact of weakening the built-in language model in RNN-T since the language model has the potential to memorize in-domain subsequence and maybe one source of the posterior skew.  We introduced noise in the reference sequence by masking words with certain probability. Specifically, during ASR training, the output symbols that were fed back into the decoder were masked with varying degrees of randomness. We used a probability of $0.2$ with no degradation in ASR performance and observed that the $\NCE$ value did not change but the $\AUC_{\NT}$ improved to $0.22$.

\begin{table}[h]
\vspace{-.1in}
\centering
\begin{tabular}{|l|c|c|c|c|c|}
\hline
\multirow{2}{*}{} & \multirow{2}{*}{$\NCE$} & \multirow{2}{*}{$\ECE$} & \multicolumn{3}{c|}{$\AUC$} \\ \cline{4-6}
 &  & & $\PRC$ & $\ROC$ & $\NT$ \\ \hline
\multicolumn{6}{|c|}{Modifications of Posteriors} \\ \hline
Raw posteriors & 0.01& 0.27	&0.95	 & 0.65 & 0.14	 \\ 
Pairwise mapping  & 0.07	&0.18	 &0.95	 & 0.65 & 0.14 \\
Temp.~scaling & 0.07	&0.12	 & 0.96 & 0.74	& 0.28 \\
LM masking & 0.01	& 0.14	 & 0.95 & 0.69	& 0.22 \\ \hline
\multicolumn{6}{|c|}{Dedicated Confidence Models} \\ \hline
Transformer & 0.40	&	0.05 & 0.99 & 0.92	&	0.61 \\ 
LSTM & 0.40	& 0.03	&	0.99 & 0.92 &	0.59 \\
AM features only & 0.36	&	0.04 & 0.99 & 0.91	&	0.54 \\ 
CRF & 		 
0.39&	0.04 &	0.99 & 0.92 & 0.59 \\ 
Word-level loss & 		 
0.38&	0.06 &	0.99 & 0.90 & 0.58 \\ 
Only Dense & 0.21	&	0.08 & 0.98 & 0.84	&	0.37 \\ 
\hline
\end{tabular}
\vspace{-.1in}
\caption{Comparison of performance of the confidence model on dictation corpus using simple modifications of posteriors and a dedicated confidence models.}
\label{confidence:dictation_model}
\vspace{-.3in}
\end{table}

\subsection{Choice of Model Architecture}
Our proposed confidence model with transformer sequence layer improves performance substantially over modifications of posteriors, achieving an $\NCE$ of $0.40$, $\ECE$ of $0.05$, and $\AUC_{\NT}$ of $0.61$ (see bottom portion of the Table~\ref{confidence:dictation_model}). For understanding the impact of different model components, we performed ablations studies by switching sequence layer from transformer layers to LSTM layers, utilizing only AM features, replacing cross-entropy with CRF loss, switching the sub-word loss to word-level loss and discarding the sequence layer altogether. 

From ablation studies, we find that the lexical features improves performance substantially, and without them $\NCE$ and $\AUC_{\NT}$ drop by 10\%. The sequence layer appears to play a critical role, discarding the sequence layer degrades performance significantly, from $\NCE$ of $0.40$ to $0.21$ and $\AUC_{\NT}$ of $0.61$ to $0.37$. The sequence layer provides contextual information that helps with the predictions of confidence scores. By tracking the performance during training, we noticed that sequence layer with LSTM overfits compared to transformers. An attention span of about 30 sub-word units appears to be sufficient to capture most of the contextual dependencies.

\begin{table*}[t]
\centering
\begin{tabular}{|l|l|c|c|c|c|c| c|c|c|c| c|} %\hline
\hline

Traffic conditions & ASR encoder & Units & $\NCE$ & $\ECE$ & $\AUC_{\ROC}$ & $\AUC_{\PRC}$ & $\AUC_{\NT}$  &
%& $\mathrm{N_{tokens}}$ & $\mathrm{TCR}$	& $\mathrm{N_{words}}$ 
$\WCR$	\\ \hline
\multirow{4}{*}{Streaming} & \multirow{2}{*}{LSTM} & Graphemes	&	0.38	&	0.02	&	0.89	&	0.95	&	0.77	&	
%4266057	&	73.82\%	&	985758	&	
73.2\%	 \\ 
& &  Morphemes	&	0.36	&	0.03	&	0.88	&	0.96	&	0.70	&
%2062858	&	78.37\%	&	898205	&
78.4\%	 \\  \cline{2-9}
& \multirow{2}{*}{Transformer} & Graphemes 	&	0.34	&	0.03	&	0.88	&	0.96	&	0.67	&	
%4598659	&	79.35\%	&	1099675	&
79.0\%	 \\  
& &  Morphemes &	0.38	&	0.03	&	0.89	&	0.96	&	0.73	&
%1945439	&	78.60\%	&	843300	&
78.4\%	 \\  \hline
\multirow{4}{*}{Non-streaming} & \multirow{2}{*}{LSTM} & Graphemes	&	0.36	&	0.02	&	0.88	&	0.96	&	0.72	&
%4704068	&	77.35\%	&	1138717	&
76.7\%	 \\ 
& & Morphemes 	&	0.30	&	0.03	&	0.86	&	0.97	&	0.59	&
%	2373420	&	82.29\%	&	1036999	&
82.5\%	 \\ \cline{2-9}
& \multirow{2}{*}{Transformer} & Graphemes	&	0.37	&	0.02	&	0.89	&	0.97	&	0.70	&	
%4771667	&	80.58\%	&	1158467	
	79.6\%	 \\ 
& & Morphemes &	0.30	&	0.02	&	0.86	&	0.97	&	0.58	&
%	2500651	&	82.97\%	&	1092893	&
83.1\%	 \\ 
\hline

\end{tabular}
\vspace{-.1in}
\caption{Confidence model performances, measured on medical conversation corpus, for diverse ASR models configurations, including different types of encoders, sub-word units, and traffic conditions.} 
\label{confidence:conversation_model}
\vspace{-.1in}
\end{table*}

\begin{figure*}[t]
    \centering
    \begin{subfigure}[b]{0.23\textwidth}
        \centering
        \includegraphics[width=\textwidth]{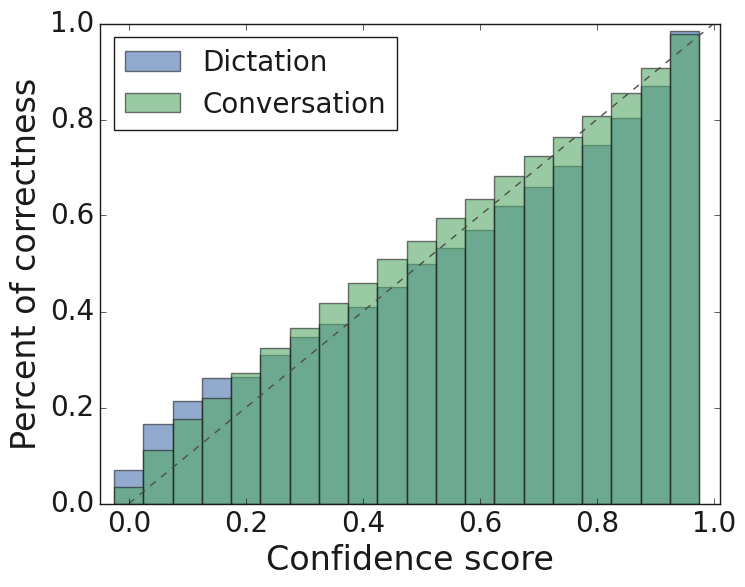}
        \caption{Calibration curve.}
        \label{fig:calibration}
    \end{subfigure}
    \hfill
    \begin{subfigure}[b]{0.23\textwidth}
        \centering
        \includegraphics[width=\textwidth]{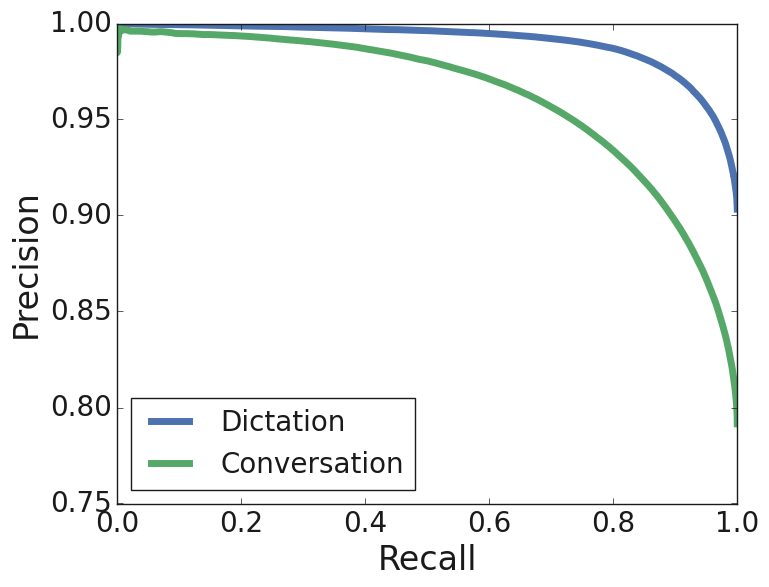}
        \caption{Precision-recall curve.}
        \label{fig:prc}
    \end{subfigure}
    \hfill
    \begin{subfigure}[b]{0.23\textwidth}
         \centering
         \includegraphics[width=\textwidth]{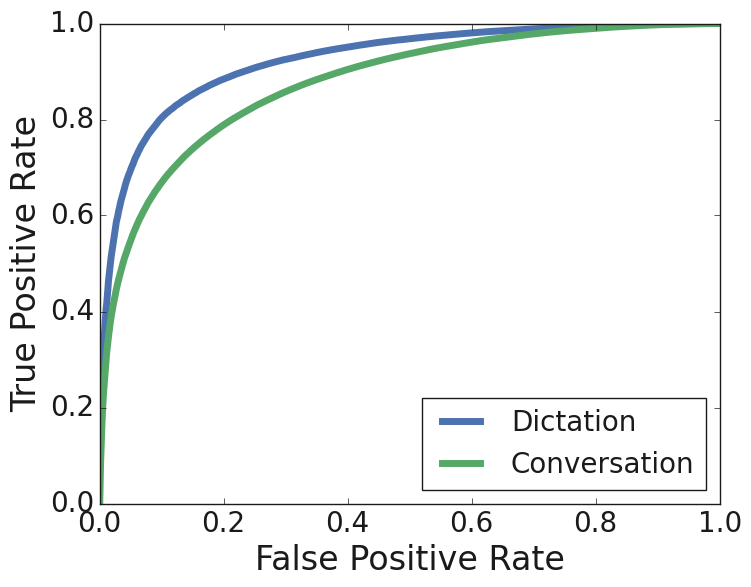}
         \caption{$\ROC$ curve.}
         \label{fig:roc}
    \end{subfigure}
    \hfill
    \begin{subfigure}[b]{0.23\textwidth}
        \centering
        \includegraphics[width=\textwidth]{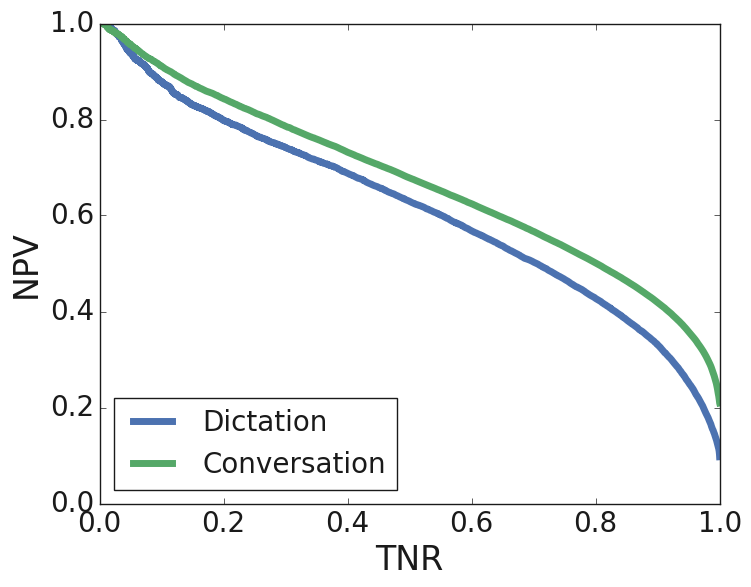}
        \caption{$\NPV\sim\TNR$ curve.}
        \label{fig:nt}
    \end{subfigure}
    \vspace{-.1in}
    \caption{Illustration of the trade-offs for picking an operating point (confidence score threshold) for downstream applications.
    }
    \label{fig:operating_point_tradeoff}
    % \vspace{-.2in}
\end{figure*}

The proposed model uses a cross-entropy loss and we evaluated the impact of replacing that with CRF loss to capture additional output dependencies. We found that the sequence layer adequately captures the necessary dependencies and the CRF layer does not help. The $\ECE$ score $0.04$ is slightly better than using cross-entropy loss $0.05$, but $\NCE$ and $\AUC_{\NT}$ are slightly worse.

One simple trick that we applied when generating confidence targets is assigning targets to all sub-words instead of masking non-ending sub-words.
This change improved $\NCE$ from $0.38$ to $0.40$ and a $\AUC_{\NT}$ from $0.58$ to $0.61$. These gains may be because the confidence model now has more examples at training time. Manual inspection of the predictions revealed that the confidence model is able to handle sub-word deletion and generalizes well on non-unique tokenization.

\subsection{Performance on diverse ASR systems}
\label{result:different_asr}

For studying the robustness of our proposed  model, we evaluated the model with eight distinct ASR model configurations which are described in~\cite{soltau2021medical}: streaming vs. non-streaming traffic conditions, LSTM vs. transformer encoder type, grapheme vs. morpheme units. These experiments were performed only on conversational corpus to limit the number of experiments and the results are reported in Table~\ref{confidence:conversation_model}. The last column enumerates the word correctness ratios ($\WCR$) of the ASR models since the performance should only be compared between systems with similar $\WCR$ measures. 

The results show that our proposed confidence model is robust to different ASR models and performs consistently without significant degradation. Comparing the performance of the confidence model with similar $\WCR$, transformer layer performs better than LSTM layer and the non-streaming condition is better than the streaming case. 

In Figure~\ref{fig:operating_point_tradeoff}, we characterize the model performance for a streaming system to illustrate the trade-off for different threshold-based operating point for post-ASR editing. Reflecting the $\ECE$ numbers, the calibration curves show an almost perfect (represented by a dotted diagonal line in Figure~\ref{fig:calibration}) alignment between the confidence scores and expected word accuracy. Similarly, the precision-recall curve and the ROC show a nice performance trade-off, mirroring the corresponding $\AUC$ metrics. However, this is overly optimistic, as illustrated by the $\NPV\sim\TNR$ curve, which shows there is still room for further improvement.

\section{Discussion}
\subsection{Effect of RNN-T's Emission Times}
\label{sec:emission_time}

In this section, we show the advantages of directly using RNN-T’s time information. One of the major differences between RNN-T~\cite{Graves2012} and LAS models~\cite{Chan2016} is that, we know the audio frame from which a token is emitted in RNN-Ts. This additional information allows the model to focus on the acoustic features around the emission times, and leads to substantially higher performance, e.g., $\NCE$ scores increasing from $0.32$ to $0.41$, and $\AUC_{\NT}$ from $0.53$ to $0.61$. It also sidesteps the need for computing cross-attention across the entire audio input as in~\cite{qiu2021learning}, which reduces the complexity from $O(UT)$ to $O(U)$ and aids faster models convergence. 

To study the impact of emission times, we compared the model performances at different cross-attention spans $k$ (see the definition of $k$ in \ref{sec:model:features}) and the results are reported in Table~\ref{confidence:dictation_model_att}. When $k$ is very large, the confidence model cross-attends to all RNN-T's acoustic frames, similar to~\cite{qiu2021learning}. At the other extreme, when $k$ is equal to $0$, the model only uses the exact frame when the token was emitted. The results show that the model achieves better performance when $k$ is smaller, reaching the highest $\NCE$, $\ECE$, and $\AUC$s at $k=1$ (three frames). This might be because the model does not have to learn the acoustic feature of a token by attending the whole audio frames. Instead, we inform the model to attend solely to the most relevant audio frames to this token. At $k=1$, the acoustic features processing can be simplified by concatenating the three frames, without the need for computing attention across the encoder output subsequence and achieves similar performance.

\begin{table}[ht]
\vspace{-.1in}
\centering
\begin{tabular}{|l|c|c|c|c|c|}
\hline
\multirow{2}{*}{$\mathrm{k}$} & \multirow{2}{*}{$\NCE$} & \multirow{2}{*}{$\ECE$} & \multicolumn{3}{c|}{$\AUC$} \\ \cline{4-6}
 &  & & $\PRC$ & $\ROC$ & $\NT$ \\ \hline
0	&	0.39	&	0.04	&	0.92	&	0.99	&	0.61	\\ \hline
1	&	0.41	&	0.04	&	0.92	&	0.99	&	0.61	\\ \hline
3	&	0.39	&	0.05	&	0.92	&	0.99	&	0.60	\\ \hline
5	&	0.38	&	0.05	&	0.92	&	0.99	&	0.59	\\ \hline
10	&	0.38	&	0.05	&	0.92	&	0.99	&	0.59	\\ \hline
15	&	0.38	&	0.04	&	0.91	&	0.99	&	0.58	\\ \hline
All	&	0.32	&	0.05	&	0.89	&	0.99	&	0.53	\\ \hline
\end{tabular}
\caption{Confidence model performances for various spans $k$ of the ASR encoder outputs for the medical dictation corpus.}
\label{confidence:dictation_model_att}
\vspace{-.1in}
\end{table}

% One potential drawback of relying on emission time with $k=1$ is that the confidence model may be sensitive to inaccurate emission times. If the emission time is off by one or two frames at inference, the performance could deteriorate substantially. Indeed, we observed this in our dictation model, where finite-state determinization of the hypotheses in the lattice led to a jitter in the emission time and the $\NCE$ nose-dived from $0.40$ to $0.01$.

% However, this can be easily addressed by introducing noise in the emission time during training. We augmented the training set with jitters in the emission time.
% For example, each token with emission time $\tau_{i}$ is augmented into five versions of training samples with emission times of $\tau_{i}-2$, $\tau_{i}-1$, $\tau_{i}$, $\tau_{i}+1$, and $\tau_{i}+2$. This data augmentation improves robustness without incurring any loss in performance, achieving an $\NCE$ score of $0.40$. 

\subsection{Challenges in Detection of Incorrect Words} \label{sec:discussion}

Our main motivations is to provide confidence scores for identifying incorrect words predicted by the ASR system. We argue that the best method to pick a threshold is using a performance trade-off curve based on True Negative Rate (TNR) vs Negative Predictive Value (NPV). From our results, we demonstrated that there is still room for improvement.

ASR systems have advanced substantially in recent years and achieve high performance with low word error rates. This leads to label imbalance in confidence data which is dominated by correctly recognized words $\truelabel$. Label imbalance not only results in the model difficult to train, but makes evaluations more challenging. $\AUC_{\PRC}$ is a commonly used metrics in prior confidence works. Due to the data imbalance, we observed that $\AUC_{\PRC}$ is consistently as good as over $0.98$ in the dictation corpus and over $0.95$ in the conversation corpus. However when picking a threshold for identifying incorrect words, it is challenging to find as many wrong predictions as possible (higher True Negative Rate ($\TNR$)) without sacrificing precision (higher Negative Predictive Value ($\NPV$)) (see Figure~\ref{fig:operating_point_tradeoff}). Hence, we argue that the choice of metrics is critically important in characterizing production systems, and $\AUC_{\NT}$ is a better metric for measuring performance of confidence scores associated with current ASR systems. We found that low $\ECE$ and high $\AUC_{\ROC}$ and $\AUC_{\PRC}$ do not necessarily guarantee a good trade-off between $\NPV$ and $\TNR$. For the current SOTA performance, as illustrated in Figure~\ref{fig:operating_point_tradeoff} $\NPV\sim\TNR$ curve, picking an operating point to detect incorrect words is still a challenging task and there is room for improvement.

Given the imbalance between correct and incorrect words, we explored different data augmentation techniques to generate more occurrences of wrong predictions. In addition, we experimented with focal loss to increase the contribution of incorrect words on the loss~\cite{lin2017focal}. 

\noindent \textbf{Data Augmentation}
We augmented the confidence training data by 3 times via decoding ASR trainings on 3 subsets of the training data. The data augmentation was found to stabilize the confidence model training and prevent overfitting on the medical dictation corpus, which is a relatively small corpus. The $\NCE$ score increased from $0.37$ to $0.4$.

\noindent \textbf{Focal Loss}
The focal loss increases the contribution of incorrect words to the loss~\cite{lin2017focal}. We set tunable parameters $\alpha = 0.15$ and $\gamma = 2$ in the loss function. Unfortunately, only marginal improvements were observed, $\AUC_{\NT}$ and $\AUC_{\ROC}$ improved from $0.61$ to $0.62$ and from $0.92$ to $0.93$ respectively. Despite these results, focal loss may be a useful regularization method for larger models with more features in the future.

\section{Conclusion and Future work}
The key contributions of our paper are below. 
(a) We propose a confidence model tailored for RNN-T models. Our model takes advantage of the emission times, which reduces the computational complexity and improves the performance. (b) We solve the non-unique tokenization and token deletion problems with a simple trick of mapping between sub-word and word sequences. Besides, this trick also amplifies the error signals and improves performance. 
(c) We show that our model is robust across eight different ASR configurations and two real-world long-form medical datasets. Unlike previous work, our models provide confidence scores not only for recognized words but also for other symbols associated with rich transcription, such as punctuation, capitalization and speaker role labels. 
(d) We illustrate the importance of $\AUC_{\NT}$ for measuring the performance of confidence model for current ASR systems where errors are disproportionately fewer than correct words. The $\AUC_{\NT}$ needs to be improved further so that downstream users can pick better operating point for identifying misrecognitions. 
(e) The proposed architecture is not restricted in speech recognition tasks, but also applicable for any other tasks using RNN-T. We will present a follow-up paper on applying RNN-T in named entity recognition and its corresponding confidence scores.

\section{Acknowledgements}
We are grateful for help and support from Ryan Yanzhang He, David Qiu, Stephen Koo, Chung-cheng Chiu, Hasim Sak, Han Lu, and David Rybach.

% \section{COPYRIGHT FORMS}
% \label{sec:copyright}

% You must include your fully completed, signed IEEE copyright release form when
% form when you submit your paper. We {\bf must} have this form before your paper
% can be published in the proceedings.

\bibliographystyle{IEEEbib}
\bibliography{paper}

\end{document}